\documentclass[twoside,11pt]{article}

\usepackage{jmlr2e}

\jmlrheading{2}{2002}{559-594}{9/01}{3/02}{Erik F. Tjong Kim Sang}
\ShortHeadings{Memory-Based Shallow Parsing}{Tjong Kim Sang}
\firstpageno{559}

\begin{document}

\title{Memory-Based Shallow Parsing}

\author{\name Erik F. Tjong Kim Sang \email erikt@uia.ua.ac.be \\
       \addr CNTS - Language Technology Group \\
             University of Antwerp\\
             Universiteitsplein 1\\
             B-2610 Wilrijk, Belgium}

\editor{James Hammerton, Miles Osborne, Susan Armstrong and
        Walter Daelemans}

\maketitle

\begin{abstract}
We present memory-based learning approaches to shallow parsing and
apply these to five tasks: base noun phrase identification, arbitrary
base phrase recognition, clause detection, noun phrase parsing and
full parsing.
We use feature selection techniques and system combination methods
for improving the performance of the memory-based learner.
Our approach is evaluated on standard data sets and the results are
compared with that of other systems.
This reveals that our approach works well for base phrase
identification while its application towards recognizing embedded
structures leaves some room for improvement.
\end{abstract}


\section{Introduction}

Memory-based learners classify data based on their similarity to
data that they have seen earlier.
They have been used for a variety of natural language
processing tasks with good results, for example for
grapheme-to-phoneme conversion \citep{hoste1999clin},
stress assignment \citep{daelemans1994cl} and
word class tagging \citep{hvh2001}.
These natural language processing tasks are classification tasks: they
require an assignment of a class to each character or to each word.
Shallow parsing is more complicated than that: it requires sequences
of words to be grouped together and be classified.

We believe that all natural language tasks can be performed successfully
by memory-based learners.
Identifying and classifying sequences of words can be converted to a
classification task by using special tag sets, for example the IOB
tag set proposed by \cite{ramshaw95}.
Parsing requires different processing levels and these can be
simulated by cascading several memory-based learners which have
been trained on different subtasks \citep{daelemans95}. 
The idea of using memory-based methods for processing natural language
has recently led to the emergence of a new paradigm: Memory-Based
Language Processing (MBLP) to which a special issue of the Journal of
Experimental \& Theoretical Artificial Intelligence was devoted
\citep{daelemans99b}.

The goal of this paper is to test the theoretic ideas about
memory-based learning applied to natural language tasks, in 
particular its application to shallow parsing.
We will implement the ideas of \cite{daelemans95}, show what problems
need to be solved, test memory-based shallow parsers and compare their
performance with those of other systems.
The tasks which we will examine are identification of base noun
phrases, recognition of phrases of arbitrary types, finding clauses,
discovering embedded noun phrases and full parsing.
Memory-based learning performs well on natural language tasks that 
require output that has relatively little structure.
In this paper we will investigate whether we can obtain equally good
results when it is applied to tasks requiring more complex outputs.

\section{Approach}

In our approach we will use three techniques.
We will use memory-based learning as base classification method for
assigning linguistic classes to data.
We will attempt to solve a weakness of this approach, disregarding
irrelevant features, by using an additional feature selection method.
Finally, we will examine the combination of several learners in order
to obtain an extra performance boost.
This section also contains information about evaluation and system
configuration for performing parameter tuning.

\subsection{Memory-Based Learning}

The basic idea behind memory-based learning is that concepts can be
classified by their similarity with previously seen concepts.
In a memory-based system, learning amounts to storing the training
data items.
The strength of such a system lies in its capability to compute the
similarity between a new data item and the training data items.
The most simple similarity metric is the overlap metric
\citep{timbl2000}. 
It compares corresponding features of the data items and adds 1 to a
similarity rate when they are different.
The similarity between two data items is represented by a number
between zero and the number of features, $n$, in which value zero
corresponds with an exact match and $n$ corresponds with two items
which share no feature value.
Here is an example:

\begin{center}
\begin{tabular}{ccccc}
TRAIN1 & man & saw & the & V\\
TRAIN2 & the & saw & .   & N\\
 TEST1 & boy & saw & the & ?\\
\end{tabular}
\end{center}

\noindent
It contains two training items of a part-of-speech (POS) tagger and
one test item for which we want to obtain a POS tag.
Each item contains three features: the word that needs to be tagged
({\it saw}) and the preceding and the next word.
In order to find the best POS tag for the test item, we compare its
features with the features of the training data items.
The test item shares two features with the first training data item
and one with the second.
The similarity value for the first training data item (1) is smaller
than that of the second (2) and therefore the overlap metric will
prefer the first.

A weakness of the overlap metric is that it regards all features as
equally valuable for computing similarity values.
Generally some features are more important than others.
For example, when we add a line ``TRAIN3 boy and the C'' to our
training data, the overlap metric will regard this new item as equally
important as the first training item.
Both the first and the third training item share two feature values
with the test item but we would like the third to receive a lower
similarity value because it does not contain the word for which we
want find a POS tag ({\it saw}).
In order to accomplish this, we assign weights to the features
in such a way that the second feature receives a higher weight than
the other two.

The method which we use to assign weights to the features is called
Gain Ratio, a normalized variant of information gain
\citep{timbl2000}.
It estimates feature weights by examining the training data and
determines for each feature how much information it contributes 
to the knowledge of the classes of the training data items.
The weights are normalized in order to account for features with
many different values.
The Gain Ratio computation of the weights is summarized in the
following formulas:

\begin{equation}
w_i = \frac{H(C)-\sum_{v\in V_i}P(v)\times H(C\mid v)}{H(V_i)}
\end{equation}

\begin{equation}
H(X) = - \sum_{x\in X} P(x){\rm log}_2P(x)
\end{equation}

\noindent
Here $w_i$ is the weight of feature $i$, $C$ the set of class
values and $V_i$ the set of values that feature $i$ can take.
$H(C)$ and $H(V_i)$ are the entropy of the sets $C$ and $V_i$
respectively and $H(C\mid v)$ is the entropy of the subset of elements
of $C$ that are associated with value $v$ of feature $i$.
$P(v)$ is the probability that feature $i$ has value $v$.
The normalization factor $H(V_i)$ was introduced to prevent that
features with low generalization capacities, like identification
codes, would obtain large weights.

The memory-based learning software which we have used in our
experiments, TiMBL \citep{timbl2000}, contains several algorithms
with different parameters.
In this paper we have restricted ourselves to using a single algorithm
(k nearest neighbor classification) with a constant parameter setting.
It would be interesting to evaluate every algorithm with all of its
parameters but this would require a lot of extra work.
We have changed only one parameter of the nearest neighbor algorithm
from its default value: the size of the nearest neighborhood region.
The learning algorithm computes the distance between the test item and
the training items.
The test item will receive the most frequent classification of the
nearest training items (nearest neighborhood size is 1).
\cite{daelemans99} show that using a larger neighborhood is harmful
for classification accuracy for three language tasks but not for
noun phrase chunking, a task which is central to this paper.
In our experiments we have found that using the three nearest
sets of data items leads to a better performance than using only 
the nearest data items.
This increase of the neighborhood size used leads to a form of
smoothing which can get rid of the influence of some data
inconsistencies and exceptions. 


\subsection{Feature Selection}
\label{sec-feat}

A disadvantage of the Gain Ratio metric used in memory-based learning
is that it computes a weight for a feature without examining other
available features.
If features are dependent, this will generally not be reflected in
their weights.
A feature that contains some information about the classification
class on its own, but none when another more informative feature is
present will receive a non-zero weight.
Features which contain little information about the classification
class will receive a small weight but a large number of them might
still overrule more important features.
These two problems will have a negative influence on the
classification accuracy, in particular when there are many features
available. 

\begin{figure}[t]
\begin{center}
\epsffile{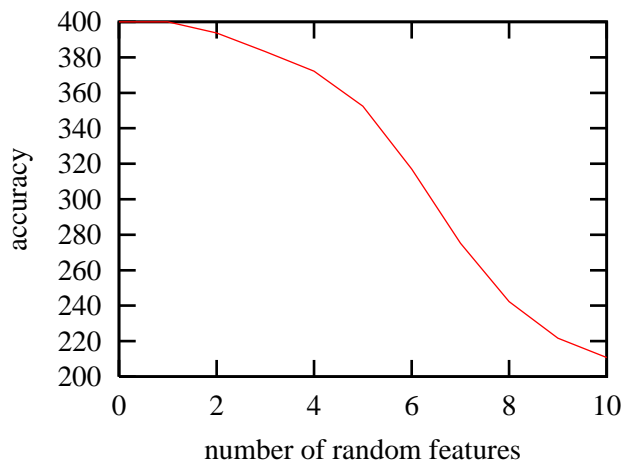}
\end{center}
\caption{Average number of correct patterns over 1000 runs of a
memory-based learner using the Gain Ratio metric for test data 
containing 400 XOR patterns after adding 0 to 10 random binary 
features.
The system performs perfectly with one random feature but when two or
more random features are added, the performance drops to about half
for 10 extra features.
}
\label{fig-feats}
\end{figure}

We have tested the capacity of Gain Ratio to deal with irrelevant
features by using it for a simple binary classification problem with
extra random features.
The problem which we chose is the XOR problem.
It contains two binary (0/1) features and a pair of these feature
values should be classified as 0 when the values are equal and as
1 when the features are different.
We have created training and test data which contained 100
examples of the four possible patterns (0/0/0, 0/1/1, 1/0/1 and
(1/1/0). 
A memory-based learner which uses Gain Ratio was able to correctly
classify all 400 patterns in the training data.
After this we added ten random binary features to both the training
data and the test data and observed the performance.
The average results of 1000 runs can be found in Figure
\ref{fig-feats}.


Without extra features the memory-based learner performs perfectly.
Adding a random feature does not harm its performance but after adding
two the system only gets 394 of the 400 patterns correct on average.
The performance drops for every extra added feature to about 211
for 10 extra features which is not much better than randomly guessing
the classes.
This small experiment shows that Gain Ratio has difficulty with 
feature sets that contain many irrelevant features.
We need an extra method for determining which features are not
necessary for obtaining a good performance.

\cite{aha94} give a good introduction to methods for selecting
relevant features for machine learning tasks.
The methods can be divided in two groups: filters and wrappers.
A filter uses an evaluation function for determining which features
could be more relevant for a classifier than others.
A wrapper finds out if one feature is more important than another by 
applying the classifier to data with either one of the features and
comparing the results.
This is requires more time than the filter approach but it generates
better feature sets because it cannot suffer from a bias difference
which may exist between the evaluation function and the classifier
\citep{john94}.

Both the filter and the wrapper method start with a set of features
and attempt to find a better set by adding or removing features and
evaluating the resulting sets.
There are two basic methods for moving through the feature space.
Forward sequential selection starts with an empty feature set and 
evaluates all sets containing one feature.
After this it selects the one with the best performance and evaluates
all sets with two features of which one is the best single feature.
Backward sequential selection starts with all features and evaluates all
sets with one feature less.
It will selects the one with the best performance and then examines
all feature sets which can be derived from this one by removing one
feature.
Both methods continue adding or removing a feature until they
cannot improve the performance.

Forward and backward sequential selection are a variant of
hill-climbing, a well-known search technique in artificial
intelligence.
As with hill-climbing, a disadvantage of these methods is that they
can get stuck in local optima, in this case a non-optimal feature set
which cannot be improved with the method used.
In order to minimize the influence of local optima, we use a combination
of the two methods when examining feature sets: bidirectional
hill-climbing \citep{caruana94}.
The idea here is to apply both adding a feature and removing a feature
at each point in the feature space.
This enables the feature selection method to backtrack from nonoptimal
choices.
In order to keep processing times down we will start with an empty
feature list just like in forward sequential selection.


\subsection{System Combination}
\label{sec-combi}

When different machine learning systems are applied to the same task,
they will make different errors.
The combined results of these systems can be used for generating an
analysis for the task that is usually better than that of any of the
participating systems, for example by choosing pattern analyses
selected by the majority of the systems.
This approach will eliminate errors that made by a minority of the
systems.
Here is a made-up example:
suppose we have five systems, c$_1$ - c$_5$, which assign binary
classes to patterns. 
Their output for eight patterns, p$_1$ - p$_8$, is as follows:

\begin{center}
\begin{tabular}{r|ccccc|l}
      & c$_1$ & c$_2$ & c$_3$ & c$_4$ & c$_5$ & correct \\\hline
p$_1$ & 0     & 0     & 0     & 0     & 0     & 0 \\
p$_2$ & 1     & 1     & 1     & 1     & 1     & 1 \\
p$_3$ & 0     & 0     & 0     & 0     & 0     & 0 \\
p$_4$ & 1     & 0     & 1     & 1     & 1     & 1 \\
p$_5$ & 0     & 0     & 1     & 0     & 0     & 0 \\
p$_6$ & 1     & 1     & 1     & 1     & 0     & 1 \\
p$_7$ & 1     & 0     & 0     & 0     & 0     & 0 \\
p$_8$ & 1     & 1     & 1     & 0     & 1     & 1 \\
\end{tabular}
\end{center}

\noindent
Each of the five systems makes an error.
We can use a combination of the five by choosing the class
that has been predicted most frequently for each pattern.
For the first three patterns this will not make a difference because
all systems predict the same class.
For pattern 4 we will choose class 1, thereby eliminating an error of
classifier 2.
Pattern 5 will be associated with class 0, thus eliminating classifier
3's only error.
Patterns 6, 7 and 8 will receive classes 1, 0 and 1 respectively,
thereby eliminating errors of classifiers 5, 1 and 4.
Thus the majority choice will generate a perfect analysis of the data.

In this paper we will evaluate different techniques for combining
system output, most of which have been put forward by
\cite{hvh2001}.
We use four voting methods and three stacked classifiers.
Voting methods assign weights to the output of the individual systems
and for each pattern choose the class with the largest accumulated
score.
The most simple voting method is the one we have used in the preceding
example: Majority Voting.
It gives all systems the same weight.
A more elaborate method is accuracy voting (TotPrecision).
It assigns a weight to each system which is equal to the accuracy of
the system on some evaluation data.

Some classes might be easier to predict than other classes and for
this reason we have also tested two voting methods which use weights
based on accuracies for particular class tags.
The first is TagPrecision.
For each output value $v$ of system $s$, it uses a weight which is
equal to the precision of that system $s$ obtained for this value $v$.
The second method is Precision-Recall.
It starts from the same weights as TagPrecision but adds to these the
probability that systems producing different output values would have
missed $v$.
For example, suppose that there are two systems $s_1$ and $s_2$, and
that for some data item $s_1$ predicts value $v_1$ while $s_2$
predicts something else. 
In that case, the probability that $s_1$ is right is
$precision(s_1,v_1)$ while the probability that $s_2$ would have
missed $v_1$ is $1-recall(s_2,v_1)$.
Precision-Recall will assign the weight
$precision(s_1,v_1)+(1-recall(s_2,v_1))$ to the event of $s_1$
predicting $v_1$. 

A stacked classifier is a classifier which processes the results of
other classifiers.
We have used three variants of stacked classifiers.
The first is called TagPair.
It examines pairs of values produced by two systems and estimates the
probability that a certain output value is associated with the pair.
In the case of the two systems $s_1$ and $s_2$ producing two distinct 
values $v_1$ and $v_2$, TagPair will examine evaluation data and find
that the value pair is associated with, for example, $v_1$ in 20\% of
the cases, $v_2$ in 70\% and $v_3$ in 10\%.
These numbers will be used as weights for the three output values and
the one that has accumulated the largest value after examining all
value pairs in the pattern, will be selected.
Unlike the voting methods, TagPair has the opportunity to choose the
correct output tag even if all systems have made an incorrect
prediction (for example, $v_3$ in this example).

The other stacked classifier which we have evaluated is the
memory-based learner itself.
We have tested it in two modes: one in which only the output of the
systems was included and one in which we included information about 
the test item.
This extra information was the word that needed to be classified, its
part-of-speech (POS) tag and the context (words/POS tags) in which it
appeared.
The memory-based learner used the same settings as described earlier
in this section: it used the Gain Ratio metric and examined a nearest
neighborhood of size three.

The weight assignment methods used by the voting methods and the
stacked classifiers suffer from the same problem as Gain Ratio:
they might fail to disregard irrelevant features.
For this reason we have often tested the combination methods both with 
all available system results as well as with a subset of these, thus
mimicking the feature selection method described earlier.
Apart from Majority Voting, all voting methods and stacked classifiers
require training data.
This means that we need both training data for the individual systems
and training data for the combinators.
We will describe how we have selected the training data in the next
section.

\subsection{Parameter Tuning}
\label{sec-partun}

In this paper, we will compare different learner set-ups and
apply the best one to standard data sets.
For example, we will examine different data representations and
test different system combination techniques.
We should be careful not to tune the system to the test data and
therefore we will only use the available training data for finding the
best configuration for the learner.
This can be done by using 10-fold cross-validation \citep{weiss91}.
The training data will be divided in ten sections of similar size and
each section will be processed by a system which has been trained on
the other nine.
The overall performance on all ten sections will be regarded as the
performance of the system.

In our experiments, we will process the data twice.
First we will let the learner generate a classification of the data.
After this the learner will process the data another time, this time
while including the classifications found earlier for the context of a
data item.
While working with n-fold cross-validation, we should be careful that
information from a test part is not accidentally used in its training
part. 
In the first processing phase we will generate classes for the first
section while using the other nine sections.
Thus information about the classes in, for example, section two is
encoded in the classes produced in section one.
If in the second phase we use the classifications of the first section 
while processing section two, we are analyzing a section while having
access to (indirect) information about the classes in the data.
Information about the classes in section two might leak to this process
via the training data, something which is undesired.

There are two ways for preventing this form of information leaking.
Both concern being more strict when it comes to creating the training 
data of the second system.
In a cascaded 10-fold cross-validation experiment, the second phase
training data for section x must be constructed without using this
section. 
This means that instead of running one 10-fold cross-validation
experiment with the first system, we need to run ten 9-fold
cross-validation experiments in order to obtain correct training data
for the ten sections in the second system.
Section one will be trained with the 9-fold cross-validation results
from sections 2-10, section 2 with 1 and 3-10 and so on.
If at any time we need to add a third phase to the cascade of systems,
we need to run 8-fold cross-validation experiments with the first
system and 9-fold cross-validation experiments with the second.
For extra systems the number of extra runs increases and the amount of
available training data for the first system decreases.

The second method for preventing training information from a
processing phase leaking to the classifications of a next phase 
is by only using results from previous phases in the test data.
In the training data we use the perfect classes rather than
the output of the previous phase.
This has two disadvantages.
First, we cannot use a feature containing the class of the focus 
word because this feature is the same as the output class.
This means that we can only use the classes of neighboring words.
Second, the opportunity to correct errors made in the first phase
will be restricted because the training data no longer contains
information about the errors made by this phase.
The advantage of this approach is that we can use all training data in
all training phases, so the problem of a diminishing quantity of
training data disappears.
This approach is especially useful with longer cascades of learners,
as for example is required in full parsing.

Here is an example to illustrate the two methods: suppose a word in
the sixth section in the second phase of a ten-fold cross-validation
experiment in chunking is represented by the following eight features: 

\begin{quote}
$w_{i-1}$ $w_i$ $w_{i+1}$ $p_{i-1}$ $p_i$ $p_{i+1}$ $c_{i-1}$ $c_{i+1}$
\end{quote}

\noindent
The goal is to find a chunk tag for word $w_i$.
The word features $w_i$, $w_{i-1}$ and $w_{i+1}$ represent, the word
itself, the preceding word and the next word, respectively.
The POS tag features $p_i$, $p_{i-1}$ and $p_{i+1}$ contain the POS
tags of the three words.
The two chunk features $c_{i-1}$ and $c_{i+1}$ hold the chunk tag of
the preceding and the next word.
The word and POS tag information have been taken from the training
data.
In the first method, the two chunk features are computed by a 
preceding phase.
If this item is part of the training data for section x, $c_{i-1}$ and
$c_{i+1}$ were generated by a nine-fold cross-validation experiment
which uses all sections except section x.
This means that the two chunk features have been generated by training
with all sections except 6 and x.
If the item is part of the test data, then the chunk features are
computed by a ten-fold cross-validation experiment (training with
sections 1-5 and 7-10).
The second method generates chunk features for the test data in the
same way but for training data it takes $c_{i-1}$ and $c_{i+1}$ from
the training data, thus preventing that they contain implicit
information about the test sections.\footnote{In case $c_{i-1}$ is part
of a previous section or $c_{i+1}$ is in a next section, they are left
empty.}

\subsection{Evaluation}
\label{sec-stat}

We will compare the results of a shallow parser with an available
hand-parsed corpus.
For this purpose we will use the precision and recall of the phrases
in the results.
Precision is the percentage of phrases found by the learner that are
correct according to the corpus.
Recall is the percentage of corpus phrases found by the learner.
It is easier to optimize a system configuration based on one
evaluation score and therefore we combine precision and recall 
in the F$_{\beta}$ rate \citep{vanrijsbergen75}:

\begin{equation}
F_{\beta} = \frac{(\beta^2+1)*precision*recall}{\beta^2*precision+recall}
\end{equation}

\noindent
$\beta$ can be used for giving precision a larger ($\beta>$1) or
smaller ($\beta<$1) weight than recall.
We do not have a preference for one or the other and therefore we use
$\beta$=1.
In previous work on shallow parsing, often a word-related accuracy
rate is used as evaluation criterion.
We do not believe that this is a good method for evaluating results of
phrase detection algorithms. 
Accuracy rates assign positive values to correctly identified
non-phrase words and to partially identified phrases.
Furthermore they will produce different numbers for the same analysis
based on the data representation used.
For these reasons, the relation between accuracy rates and F$_{\beta}$
rates is poor and preference should be given to using the latter.

Accuracy rates have one advantage over F$_{\beta}$ rates: standard
statistical tests can be used for determining if the difference
between two accuracy rates is significant.
Accuracy is a relatively simple function $correct/processed$ where
$processed$ is the number of items that have been processed and
$correct$ is the number of items that received the correct class.
Unfortunately, F$_{\beta=1}$ is more complex: after some arithmetic 
we get $2*correct/(found+corpus)$ where $found$ is the number of
phrases found by the learner, $correct$ the number of phrases found
that were correct and $corpus$ the number of phrases in the corpus
according to some gold standard.
The value of the $corpus$ variable is an upper bound on the variable
$correct$. 
The complexity of the F$_{\beta=1}$ computation makes it hard to
apply standard statistical tests to F$_{\beta=1}$ rates.

\cite{yeh2000} offers a method for computing significance values for
F$_{\beta=1}$ rate comparisons: by using computationally-intensive
randomization tests.
His approach requires test data classifications for all systems that
need to the compared.
Usually we only have access to the test data classifications of our
own system and therefore we have used a variant of these 
randomization tests presented: bootstrap resampling
\citep{noreen89}.
The basic idea of this approach is to regard the test data
classifications as a population of cases.
A random sample of this population can be created by arbitrarily
choosing cases with replacement.
We can create many random samples of the same size as the test data
and compute an average F$_{\beta=1}$ rate over the samples and a
standard deviation for this average.
These statistical measures can be used for deciding if the performance
of another system is significantly different from our system.
Since we do not know if the performance of our system is distributed
according to a normal distribution, we will determine
significance boundaries in such a way that 5\% of the samples evaluate
worse (or better) than the chosen boundary.

\section{Chunking}

In this section we will apply a memory-based learner to chunking,
identifying base phrases.
The section starts with a some background information on this task.
After this we will present the results of our experiments with base
noun phrase identification and our work targeted at finding base
phrases of arbitrary types.

\subsection{Task Overview}
\label{sec-chunkov}

A text chunker divides sentences in phrases which consist of a
sequence of consecutive words which are syntactically related.
The phrases are nonoverlapping and nonrecursive.
In the beginning of the nineties, \cite{abney91} suggested to use
chunking as a preprocessing step of a parser.
Ten years later, most statistical parsers contained a chunking phase
(for example \cite{ratnaparkhi98}).
In this study, we will divide chunking in two subtasks: finding only
noun phrases and identifying arbitrary chunks.

Machine learning approaches towards noun phrase chunking started with 
work by \cite{church88} who used bracket frequencies associated with
POS tags for finding noun phrase boundaries in text.
In an influential paper about chunking, \cite{ramshaw95} show that
chunking can be regarded as a tagging task.
Even more importantly, the authors propose a training and test data
set that are still being used for comparing different text chunking
methods.
These data sets were extracted from the Wall Street Journal part of
the Penn Treebank II corpus \citep{marcus93}.
Sections 15-18 are used as training data and section 20 as test
data.\footnote{The noun phrase identification data is available from
{\tt ftp://ftp.cis.upenn.edu/pub/chunker/}}
In principle, the noun phrase chunks present in the material are noun
phrases that do not include other noun phrases, with initial material
(determiners, adjectives, etc.) up to the head but without
postmodifying phrases (prepositional phrases or clauses)
\citep{ramshaw95}.

The noun phrase chunking data produced by \cite{ramshaw95} contains a
couple of nontrivial features.
First, unlike in the Penn Treebank, possessives between two noun
phrases have been attached to the second noun phrase rather than the
first.
An example in which round brackets mark chunk boundaries: {\it ( Nigel
Lawson ) ('s restated commitment )}: the possessive {\it 's} has been
moved from {\it Nigel Lawson} to {\it restated commitment}.
Second, Treebank annotation may result in nonexpected noun phrase
annotations: {\it British Chancellor of ( the Exchequer ) Nigel
Lawson} in which only one noun chunk has been marked. 
The problem here is that neither {\it British Chancellor} nor {\it
Nigel Lawson} has been annotated as separate noun phrases in the
Treebank.
Both {\it British ... Exchequer} and {\it British ... Lawson} are
annotated as noun phrases in the Treebank but these phrases could not
be used as noun chunks because they contain the smaller noun phrase
{\it the Exchequer}.

\cite{ramshaw95} proposed to encode chunks with tags: I for words that
are inside a noun chunk and O for words that are outside a chunk.
In case one noun phrase immediately follows another one, they
use the tag B for the first word of the second phrase in order to show
that a new phrase starts there.
With the three tags I, O and B any chunk structure can be encoded.
This representation has two advantages. 
First, it enables trainable POS taggers to be used as chunkers by
simply changing their training data.
Second, it minimizes consistency errors which appear with the bracket
representation where open and close brackets generated by the learner
may not be balanced.
Here is an example sentence first with noun phrases encoded by pairs of
brackets and then with the Ramshaw and Marcus IOB representation:

\begin{quote}
In ( early trading ) in ( Hong Kong ) ( Monday ) , ( gold ) was quoted \\
at ( \$ 366.50 ) ( an ounce ) .

In$_O$ early$_I$ trading$_I$ in$_O$ Hong$_I$ Kong$_I$ Monday$_B$ ,$_O$ gold$_I$ was$_O$ quoted$_O$ \\
at$_O$ \$$_I$ 366.50$_I$ an$_B$ ounce$_I$ .$_O$ 
\end{quote}

\noindent 
\cite{tksveenstra99eacl} presents three variants on the Ramshaw and
Marcus representation and shows that the bracket representation can
also be regarded as a tagging representation with two streams of
brackets.
They named the variants IOB2, IOE1 and IOE2 and used IOB1 as name for
the Ramshaw and Marcus representation.
IOB2 was the same as IOB1 but now every chunk-initial word receives tag B.
IOE1 differs from IOB1 in the fact that rather than the tag B, a tag E
is used for the final word of a noun chunk which is immediately
followed by another chunk.
IOE2 is a variant of IOE1 in which each final word of a noun phrase is
tagged with E.
The bracket representations use open brackets for phrase-initial
words, close brackets for phrase-final words and a period for all
other words.
Table \ref{tab-repr} contains example tag sequences for all six tag
sequences for the example sentence.

\begin{table}[t]
\begin{center}
\begin{tabular}{|c|ccccccccccccccccc|}\hline
IOB1 &  O&I&I&O&I&I&B&O&I&O&O&O&I&I&B&I&O \\
IOB2 &  O&B&I&O&B&I&B&O&B&O&O&O&B&I&B&I&O \\
IOE1 &  O&I&I&O&I&E&I&O&I&O&O&O&I&E&I&I&O \\
IOE2 &  O&I&E&O&I&E&E&O&E&O&O&O&I&E&I&E&O \\
O    &  .&$[$&.&.&$[$&.&$[$&.&$[$&.&.&.&$[$&.&$[$&.&. \\
C    &  .&.&$]$&.&.&$]$&$]$&.&$]$&.&.&.&.&$]$&.&$]$&. \\\hline
\end{tabular}
\end{center}
\caption{The chunk tag sequences for the example sentence
{\it In early trading in Hong Kong Monday , gold was quoted at \$
366.50 an ounce . } 
for six different tagging formats.
The {\tt I} tag has been used for words inside a chunk, {\tt O}
for words outside a chunk, {\tt B} and {\tt [} for 
chunk-initial words and {\tt E}, {\tt ]} for chunk-final words and
periods for words that are neither chunk-initial nor chunk-final.
}
\label{tab-repr}
\end{table}

The representation variants are interesting because a learner will
make different errors when trained with data encoded in a different
representation.
This means that we can train one learner with five\footnote{The
combination of open and close brackets, O+C, will be regarded as one
data representation.} 
data representations and obtain five different analyses of the data
which we can combine with system combination techniques.
Thus the different data representations may enable us to improve the
performance of the chunker.
The data representations can be used both for noun phrase chunking
and for arbitrary chunking.
In the latter task, more than one chunk type exists so the tags need
to be expanded with type-specific suffixes.
For example: B-VP, I-VP, E-VP, $[$-VP and $]$-VP.

The arbitrary chunking task was more difficult to design because many
interesting phrase types often contain parts which belong to other
phrases \citep{tksbuchholz2000conll}.
For example, verb phrases may contain noun phrases and prepositional
phrases often include a noun phrase.
Furthermore, noun phrases may contain quantitative or adjective phrases
which may prevent them from being identified as noun chunks.
The noun, verb and prepositional phrases should be included and
therefore the following measures have been taken when constructing the
data for the arbitrary chunking task:
First, a couple of phrase types, for example quantifier phrases and
adjective phrases, have been removed from places where they prevented
the identification of noun phrases.
This made possible annotating more phrases as noun chunks.
Second, some phrase types in the annotated data, for example verb
phrases and prepositional phrases, lack material that has already been
included in a phrase of another type.
Third, adjacent verb clusters have been put in one flat verb phrase
unlike in the Treebank where often each verb starts a new phrase.
And fourth, adverbial phrase boundaries have been removed from
adjective phrases and verb phrases to allow all material to be
included in the mother phrase. 

This chunk definition scheme will generate data in which most of the
tokens have been assigned to a chunk of some type.
The odd tokens that fall out are usually punctuation signs.
This chunk scheme has been used for generating training and test data
for the CoNLL-2000 shared task \citep{tksbuchholz2000conll}.
The data contains the same segments of the Wall Street Journal part of
the Penn Treebank as the noun phrase data of \cite{ramshaw95}: sections
15-18 as training data and section 20 as test data.\footnote{The
CoNLL-2000 shared task data is available from
{\tt http://lcg-www.uia.ac.be/conll2000/chunking/}}
We will use these data sets in our arbitrary chunking experiments.

The training and the test data contain two types of features: words
and POS tags.
The words have been taken from the Penn Treebank.
The POS tags of the Treebank have been manually checked and therefore
they should not be used in the chunking data.
In future applications, the chunking process will be applied to a text
with POS tags that have been generated automatically.
These POS tags will contain errors and therefore the performance of
the chunker will be worse than when applied to a Treebank text with
manually checked POS tags.
If we want to obtain realistic performance rates, we should work with
automatically generated POS tags in our shallow parsing experiment.
Conform with earlier work like that of \cite{ramshaw95}, we have used
POS tags that were generated by the Brill tagger \citep{brill94}.

\subsection{Noun Phrase Recognition}

We will use a memory-based learner to find noun phrase chunks in text.
In order to determine the best configuration for the learner, we will
test different system configurations on the standard training data
sets put forward by \cite{ramshaw95}.
We will evaluate different feature sets for representing words.
Additionally, we will use the five data representations for generating
different system results and use system combination techniques for
combining these results.

In our experiments we will represent words as sets of words and POS
tags.
These sets contain the word itself, its part-of-speech (POS) tag and a
left and right context of a maximum of four words and POS tags on each
side, 18 features in total.
We have explained in Section \ref{sec-feat} that memory-based learners
equipped with the Gain Ratio metric have difficulty in dealing with
irrelevant features.
Therefore we will use a feature selection method, bi-directional
hill-climbing starting with zero features, for finding the best
subset of the 18 features for each different data representation.

The memory-based learner will make two passes over the data.
First, it will attempt to predict the noun phrases in the data as well
as possible.
After this it will use the output of this first pass as information
about the noun phrases in the immediate context of the current word.
This means that the second pass has access to the 18 features of the
first pass plus the chunk tags of the two words immediately in front
of the current word and the chunk tags of the two words immediately
following the current word.
This cascaded approach was chosen because it was useful for improving
overall performance in our earlier work \citep{tksveenstra99eacl}.
We omitted the chunk tag for the current word because including it
gave a negative bias to the chunker performance.
Gain Ratio would correctly identify it as a feature which contained a
lot of information about the output class and the weight it assigned
to it would make it hard for the other features to influence the
output class at all \citep{tksveenstra99eacl}.\footnote{
The problem of using the predicted class of the current word was a
result of an earlier study in which we did not use feature selection.
The selection method used in this study would probably have disregarded
this feature automatically.
It would start out as the most informative feature but with the
feature on its own we would get a worse performance than with
combinations of other features (we perform feature selection while
keeping the five best combinations).
}

We performed a cascaded feature search while using five different data
representations on the training data of \cite{ramshaw95} in a 10-fold
cross-validation approach.
We prevented information leaking in the second phase conform Section
\ref{sec-partun}  by using the estimated chunk tags for test data and 
using the corpus tags in the training data. 
In this way we made sure that when the test data consisted of section
x, no information about section x was available in the training data. 
The results of the 10-fold cross-validation experiments can be found
in Table \ref{tab-np10a}.
In the best feature sets of the first pass most of the nine POS tag
features are used (almost eight on average) but interestingly enough
only a few of the word features (just over four on average).
The best sets for the second pass use fewer POS tag features (under
seven), fewer word tags (just over two) and most of the chunk features
(about three).
The table shows that a wide context is more important for the POS
features than for the chunk features and less important for the word
features. 

\begin{table}[t]
\begin{center}
\begin{tabular}{|l|c|ll|c|lll|}\cline{2-8}
\multicolumn{1}{l|}{train} & \multicolumn{3}{c|}{Pass 1}  
                           & \multicolumn{4}{c|}{Pass 2} \\\hline
Repr.& F$_{\beta=1}$ & \multicolumn{2}{c|}{features} & 
       F$_{\beta=1}$ & \multicolumn{3}{c|}{features} \\\hline
IOB1 & 91.88 & word$_{-4..0}$ & POS$_{-2..3}$
     & 92.54 & word$_{-2..0}$ & POS$_{-4..3}$ & chunk$_{-2,-1,1,2}$\\
IOB2 & 91.78 & word$_{-1..0}$ & POS$_{-4..3}$
     & 92.29 & word$_{-1..0}$ & POS$_{-4..2}$ & chunk$_{-1,1,2}$\\
IOE1 & 91.64 & word$_{0..1}$  & POS$_{-3..3}$
     & 92.28 & word$_{0..1}$  & POS$_{-3..3}$ & chunk$_{-1,1,2}$\\
IOE2 & 92.19 & word$_{-3..4}$ & POS$_{-4..4}$
     & 92.59 & word$_{0..1}$  & POS$_{-1..3}$ & chunk$_{-2,-1,1,2}$\\
O    & 96.04 & word$_{-2..0}$ & POS$_{-4..3}$
     & 96.11 & word$_{-1,0}$  & POS$_{-4..1}$ & chunk$_{-1,2}$\\
C    & 96.43 & word$_{0..4}$  & POS$_{-4..4}$ 
     & 96.45 & word$_{0..2}$  & POS$_{-4..2}$ & chunk$_{-2,-1,1}$\\\hline
\end{tabular}
\caption{Best F$_{\beta=1}$ found for six data representations in two
passes while using a bi-directional hill-climbing feature search
algorithm in a 10-fold cross-validation process applied to the
training data for the noun phrase chunking task.
Note that the rates obtained for the O (open bracket) and C (close
bracket) representations are for phrase starts and phrase ends
respectively and thus higher than for the first four which evaluate
complete phrase identification.
}
\label{tab-np10a}
\end{center}
\end{table}

Our motive for processing six representations rather than one was to
obtain different results which we could combine in order to improve
performance.
System combination can be seen as a second cascade behind passes one
and two.
For reasons mentioned in Section \ref{sec-partun}, adding a second
cascade in a 10-fold cross-validation experiment requires taking extra
care to prevent information leaking from a training data at one level
to the training data of the next level.
We have taken care of this problem by preparing the training data of
the combination techniques with 9-fold cross-validation runs which
were independent of the 10-fold cross-validation experiments used
for generating the test data.
For example, the test data for the first section was generated by
training with sections 2-10 twice, first without information about
context chunk tags and then with the perfect information of the
context chunk tags. 
The training data was generated with a 9-fold cross-validation process
on sections 2-10, also first without context chunk tags and then with
perfect context chunk tags.
By working this way it was impossible for information about the first
section to enter the training data of the combination processes.

Most system combination techniques require results that are in the
same format.
We have results in six different formats which means that we need to
convert them to one format.
Since we do not know which of the formats would suit the combination
process best, we have evaluated all formats.
The four IO formats can trivially be converted to each other and to
the O and the C format.
The conversion of the two bracket formats to the other four is
nontrivial.
The two data streams have been generated independently of each other
and this means that they may contain inconsistencies.
We have chosen to get rid of these by removing all brackets which
cannot be matched with the closest candidate.
For example, if we have a structure like {\it ( a ( b c ) d )} then
the first bracket will be removed because it cannot be matched with
the second bracket.
The second and third will be kept because they match.
Finally, the fourth will be removed because it cannot be matched with
the third.
We obtain the balanced structure {\it a ( b c ) d} which can trivially
be converted to the four IO formats.

\begin{table}[t]
\begin{center}
\begin{tabular}{|l|c|c|c|c|c|}\cline{2-6}
\multicolumn{1}{l|}{train}
                 &  IOB1 &  IOB2 &  IOE1 &  IOE2 &  O+C\\\hline
{\bf all systems} &      &       &       &       &       \\
Majority         & 93.06 & 93.06 & 93.14 & 93.12 & 93.35 \\
TotPrecision     & 93.06 & 93.05 & 93.13 & 93.05 & 93.35 \\
TagPrecision     & 93.06 & 93.10 & 93.11 & 93.11 & 93.35 \\
Precision-Recall & 93.06 & 93.10 & 93.11 & 93.08 & 93.35 \\
TagPair          & 93.05 & 93.14 & 93.10 & 93.13 & 93.36 \\
MBL              & 93.14 & 93.12 & 93.07 & 92.92 & 93.35 \\
MBL+             & 92.81 & 92.74 & 92.91 & 92.78 & 93.29 \\\hline
{\bf some systems} &     &       &       &       &       \\
Majority         & 93.02 & 93.12 & 93.08 & 92.99 & 93.37 \\
TotPrecision     & 93.02 & 93.12 & 93.08 & 92.99 & 93.37 \\
TagPrecision     & 93.04 & 93.13 & 93.10 & 92.99 & 93.37 \\
Precision-Recall & 93.04 & 93.13 & 93.13 & 93.04 & 93.37 \\
TagPair          & 93.08 & 93.16 & 93.12 & 93.05 & 93.37 \\
MBL              & 93.12 & 93.18 & 93.18 & 93.03 & 93.38 \\\hline
\end{tabular}
\caption{
F$_{\beta=1}$ rates obtained on 10-fold cross-validation experiments
on the noun phrase chunking data while combining results obtained with
five different data representations.
All five representations have been tested and best rates have been
obtained while using the the combined bracket representation O+C.
All combination results are better than any result of the individual
systems (92.59, see Table \ref{tab-np10a}) and generally combing five
systems led to better results than when only three or four were used.
The best results have been obtained with a stacked memory-based
classifier that used all system results except those generated with
IOE1.
However, the performance differences are small.
}
\label{tab-np10b}
\end{center}
\end{table}


We have combined the five results of pass two of the 10-fold
cross-validation experiments on the noun phrase chunking training
data (O and C have now been regarded as one data stream O+C).
We have used the system combination techniques described in Section
\ref{sec-combi}: Majority Voting, TotPrecision, TagPrecision,
Precision-Recall, TagPair and two variants of a stacked memory-based
learner.
The first stacked learner did not use any context information while
the second one had access to a limited amount of context information:
the current word, the current POS tag or pairs containing the current
POS tag and one of the three current word, previous POS tag or next
POS tag.
We have performed combination experiments with all five data streams
and with all subsets of three and four data streams.
The results can be found in Table \ref{tab-np10b}.
For the second stacked classifier we only included the best results
(obtained with context feature current POS tag).
System combination improved performance: the worst result of the
combination techniques is still better than the best result of the
individual systems.
The differences between the combination techniques are small.
Furthermore, system combination with the four IO data representations
leads to similar results but the combined bracket representation
consistently obtains higher F$_{\beta=1}$ rates.
It should be noted though that while combination of the data with the
IO representations leads to similar precision and recall figures, O+C
obtains its higher F$_{\beta=1}$ rates with high precision rates and
lower recall rates.

Since the performance differences between the combination techniques
displayed in Table \ref{tab-np10b} are small, we are relatively free
in selecting a technique for further processing.
We chose Majority Voting because it is the simplest of the
combination techniques that were tested since it does not
require extra combinator training data like the other techniques.
It does seem reasonable to use the O+C representation during the
combination process because the best results have been obtained with
this representation.
We will restrict ourselves to a few systems rather than combining all
because Majority Vote in combination with the O+C representation
obtained a slightly higher F$_{\beta=1}$ rate that way.
The best rate was obtained while using only the systems with data
representations IOB1, IOE2 and O+C so we restrict ourselves to
these three.
This leaves us with the following processing scheme:

\begin{enumerate}
\item Process the test data with a memory-based model generated from
      the training data.
      Use the features shown in Table \ref{tab-np10a} (Pass 1) and
      generate output data streams while using the representations
      IOB1, IOE2, O and C.
\item Perform a second pass over the test data with another
      memory-based model obtained from the training data.
      Again use the features shown in Table \ref{tab-np10a} (Pass 2).
      In the test data, use the estimated chunk tags from the previous
      run as chunk tag features and in the training data use the
      corpus chunk tags as chunk features.
      Perform these passes four times, once for each of the data
      representations IOB1, IOE2, O and C.
\item Convert the output for the data representations IOB1 and IOE2 to
      the O and the C format.
\item Combine the three O data streams (IOB1, IOE2 and O) with
      Majority Voting and do the same for the three C data streams
      (IOB1, IOE2 and C).
\item Remove brackets from the resulting O and C data streams which
      cannot be matched with other brackets. 
      The balanced bracket structure is the analysis of the test data
      that is the output of the complete system.
\end{enumerate}

\noindent
We have applied this procedure to the data sets of \cite{ramshaw95}:
sections 15-18 of the Wall Street Journal part of the Penn Treebank
\citep{marcus93} as training data and section 20 of the same corpus
as test data.
The system obtained a F$_{\beta=1}$ rate of 93.34 (precision 94.01\%
and recall 92.67\%).
This is a modest improvement of our earlier work \citep{tks2000naacl}
in which we did not use feature selection and where we obtained an 
F$_{\beta=1}$ rate of 93.26.
In order to estimate significance thresholds, we have applied a
bootstrap resampling test to the output of our system.
We created 10,000 populations by randomly drawing sentences with
replacement from the system results.
The number of sentences in each population was the same as in the 
test corpus. 
The average F$_{\beta=1}$ of the 10,000 populations was 93.33 with 
a standard deviation of 0.24.
For 5 percent of the populations, the F$_{\beta=1}$ rate was equal to
or lower than 92.93 and for another 5 percent it was equal to or higher
than 93.73.
Since 93.26 is between the two significance boundaries, our current
system does not perform significantly better than the previous version
without feature selection.

\subsection{Arbitrary Phrase Identification}
\label{sec-chuarb}

Our work with chunks of arbitrary types\footnote{The results of our
arbitrary phrase identification work have earlier been presented by
\cite{tks2000conll}.}
is similar to that with noun phrase chunks apart from two facts.
First, we refrained from using feature selection methods.
Applying these methods did not gain us much for noun phrase
chunking but they required a lot of extra computational work.
Therefore we went back to using a fixed set of features in these
experiments.
The context size we used here was four left and four right for words
and POS tags in the first pass over the data, and three left and three
right for words and POS tags, and two left and two right without the
focus for chunk tags in the second pass.
This means that both first and second pass use 18 features.
The second pass has only been used for the four IO data
representations.
Table \ref{tab-np10a} shows that the second pass improved the
performance of the first pass only by a small margin for the two
bracket representations O and C.

The second difference between this study and the one for noun phrase
chunks originates from the fact that apart from chunk boundaries, we
need to find chunk types as well.
We can approach this task in two ways.
First, we could train the learner to identify both chunk boundaries and
chunk types at the same time.
We have called this approach the Single-Phase Approach.
Second, we could split the task and train a learner to identify all
chunk boundaries and feed its output to another classifier which
identifies the types of the chunks (Double-Phase Approach).
A computationally-intensive approach would be to develop learners for
each different chunk type.
They could identify chunks independently of each other and words
assigned to more than one chunk could be disambiguated by choosing the
chunk type that occurs most frequently in the training data (N-Phase
Approach). 
Since we did not know in advance which of these three processing
strategies would generate the best results, we have evaluated all
three.

In order to find the best processing strategy and the best combination
technique, we have performed several 10-fold cross-validation
experiments on the training data.
We have processed this data for each processing strategy and in each
of the six data representations earlier used for noun phrase chunking.
After this we have used the seven combination techniques presented in
Section \ref{sec-combi} for combining these.
The results can be found in Table \ref{tab-xp10}.
Of the three processing strategies, the N-Phase Approach generally
performed best with Double-Phase being second best and Single-Phase
performing worst.
Again, system combination improved all individual results.
There were only small differences between the seven combination
techniques when compared for the same processing approach.
The only exception were the two stacked MBL classifiers applied to the
Single-Phase Approach results.
They did about 0.3 F$_{\beta=1}$ rate better than most of the other
combination techniques.

\begin{table}[t]
\begin{center}
\begin{tabular}{|l|c|c|c|}\cline{2-4}
\multicolumn{1}{l|}{train} & SP & DP & NP \\\hline
IOB1             & 90.68 & 91.59 & 92.02 \\
IOB2             & 90.77 & 91.65 & 91.94 \\
IOE1             & 90.94 & 91.60 & 91.90 \\
IOE2             & 91.21 & 91.97 & 91.99 \\
O+C              & 91.57 & 91.97 & 91.51 \\\hline
Majority         & 91.96 & 92.34 & 92.62 \\
TotPrecision     & 91.97 & 92.34 & 92.62 \\
TagPrecision     & 91.98 & 92.34 & 92.62 \\
Precision-Recall & 91.96 & 92.34 & 92.62 \\
TagPair          & 92.08 & 92.34 & 92.65 \\
MBL              & 92.32 & 92.35 & 92.75 \\
MBL+             & 92.40 & 92.32 & 92.72 \\\hline
\end{tabular}
\end{center}
\caption{F$_{\beta=1}$ rates obtained for the three processing
strategies, Single-Phase Approach (SP), Double-Phase Approach (DP) and
N-Phase approach (NP), when applied to the training data of the
CoNLL-2000 shared task (arbitrary chunking) while using five different
data representations and seven system combination techniques.
In all cases, system combination led to performances that were better
than the individual system results.
The computationally-intensive N-Phase Approach does better than the
other two.
}
\label{tab-xp10}
\end{table}


The best result was generated with the N-Phase Approach in combination
with a stacked memory-based classifier (MBL, 92.76).
A bootstrap resampling test with 8000 random populations generated the
90\% significance interval 92.60-92.90 which means that this result
is significantly better than any Single-Phase or Double-Phase result.
However, the N-Phase approach has a big computing overhead: the
number of passes over the data is at least N times the number of
representations.
Therefore, we have chosen the Double-Phase Approach combined with
Majority Voting for our further work.
This approach combines a reasonable performance with computational
efficiency.
The Single-Phase Approach is potentially faster but its performance
is worse unless we use a stacked classifier which requires extra
combinator training data.

When we applied the Double-Phase Approach combined with Majority
Voting to the CoNLL-2000 data sets, we obtained an F$_{\beta=1}$ rate
of 92.50 (precision 94.04\% and recall 91.00\%).
An overview of the performance rates of the different chunk types can
be found in Table \ref{tab-xp}.
Our system does well for the three most frequently occurring chunk
types, noun phrases, prepositional phrases and verb phrases, and less
well for the other seven.
The chunk type UCP which occurred in the training data, was not
present in the test data.
With this result, our memory-based arbitrary chunker finished third of
eleven participants in the CoNLL-2000 shared task.
The two systems that performed better were Support Vector Machines 
\citep[][F$_{\beta=1}$=93.48]{kudoh2000} and Weighted Probability
Distribution Voting \citep[][F$_{\beta=1}$=93.32]{hvh2000}.

\begin{table}[t]
\begin{center}
\begin{tabular}{|l|c|c|c|}\cline{2-4}
\multicolumn{1}{l|}{test data}
                 & precision & recall & F$_{\beta=1}$ \\\hline
ADJP  & 85.25\% & 59.36\% & 69.99 \\
ADVP  & 85.03\% & 71.48\% & 77.67 \\
CONJP & 42.86\% & 33.33\% & 37.50 \\
INTJ  &100.00\% & 50.00\% & 66.67 \\
LST   &  0.00\% &  0.00\% &  0.00 \\
NP    & 94.14\% & 92.34\% & 93.23 \\
PP    & 96.45\% & 96.59\% & 96.52 \\
PRT   & 79.49\% & 58.49\% & 67.39 \\
SBAR  & 89.81\% & 72.52\% & 80.25 \\
VP    & 93.97\% & 91.35\% & 92.64 \\\hline
all   & 94.04\% & 91.00\% & 92.50 \\\hline
\end{tabular}
\end{center}
\caption{
The results per chunk type of processing the test data with the
Double Pass Approach and Majority Voting.
Although the data is formatted differently than the noun phrase
chunking data, the NP F$_{\beta=1}$ rate here (93.23) is close to 
that of our NP chunking F$_{\beta=1}$ rate (93.34).
} 
\label{tab-xp}
\end{table}


\section{Parsing}

In this section we will examine the application of memory-based
shallow parsing to generating embedded structures.
We will examine three tasks: clause identification, noun phrase
parsing and full parsing.
Whenever possible, we will use the methods that we have applied to
chunking in the previous section.

\subsection{Clause Identification}
\label{sec-clauses}

In clause identification the goal is to divide sentences in clauses
which typically contain a subject and a predicate.
We have used the clause data of the CoNLL-2001 shared task
\citep{tksdejean2001conll} which was derived from the Wall Street
Journal Part of the Penn Treebank \citep{marcus93}.
Here is an example sentence from the Treebank, with all information
but words and clause brackets omitted:

\begin{quote}
\noindent
(S Coach them in\\
\hspace*{0.25cm}(S--NOM handling complaints)\\
\hspace*{0.25cm}(SBAR--PRP so that\\
\hspace*{0.50cm}(S they can resolve problems immediately)\\
\hspace*{0.25cm})\\
\hspace*{0.25cm}.\\
)
\end{quote}

\noindent
This sentence contains four clauses.
In the data that we have worked with, the function and type
information has been removed.
This means that the type tags NOM and PRP have been omitted and that
the SBAR tag has been replaced by S.
Like the chunking data, these data sets contained words and
part-of-speech tags which were generated by the Brill tagger
\citep{brill94}. 
Additionally they contained chunk tags which were computed by the
arbitrary chunking method we discussed in the previous section.

We have approached identifying clauses in the following
way:\footnote{This approach and the results achieved with it have
earlier been discussed by \cite{tks2001conll}.}
first we evaluated different memory-based learners for predicting the
positions of open clause brackets and close clause brackets,
regardless of their level of embedding.
The two resulting bracket streams will be inconsistent and in order to
solve this we have developed a list of rules which change a possibly
inconsistent set of brackets to a balanced structure.
The evaluation of the learners and the development of the balancing
rules will be done with 10-fold cross-validation of the CoNLL-2001
training data.
Information leaking is prevented by using corpus clause tags as
context features in the training data of cascaded learners rather than
clause tags computed in a previous learning phase.
The best learner configurations and balancing rules found will be
applied to the data for the clause identification shared task.

Like in our noun phrase chunking work, we have tested memory-based
learners with different sets of features.
At the time we performed these experiments, we did not have access to
feature selection methods and therefore we have only evaluated a few 
fixed feature configurations:

\begin{enumerate}
\itemsep -0.1cm
\item words only (w)
\item POS tags only (p)
\item chunk tags only (c)
\item words and POS tags (wp)
\item words and chunk tags (wc)
\item POS tags and clause tags (pc)
\item words, POS tags and chunk tags (wpc)
\end{enumerate}

\noindent
All feature groups were tested with four context sizes: no context
information or information about a symmetrical window of one, two or
three words.
Like in our chunking work, we want to check if an improved performance
can be obtained by using system combination.
However, since we attempt to predict brackets at all levels in one
step, we cannot use the five data representations here.
Instead we have evaluated combination of some of the feature
configurations mentioned above: a majority vote of the three using a
single type of information (1+2+3), a majority vote of the three using
pairs of information (4+5+6) and a majority vote of the previous two
and the one using three types of information (7+(1+2+3)+(4+5+6)).
The last one is a combination of three results of which two themselves
are combinations of three results.

Clauses may contain many words and it is possible that the maximal
context used by the learner, three words left and right, is not enough
for predicting clause boundaries accurately.
However, we cannot make the context size much larger than three
because that would make it harder for the learner to generalize.
We have tried to deal with this problem by evaluating another set of
features which contain summaries of sentences rather than every word.
Since we have chunk information of the sentences available, we
can compress them by removing all words from each chunk except the
main one, the head word.
The head words can be generated by a set of rules put forward by
\cite{magerman95} and modified by \cite{collins99}.\footnote{Available
on http://www.research.att.com/\~{ }mcollins/papers/heads}
After removing the nonhead words from each chunk, we can replace the
POS tag of the remaining word with the chunk tag and thus obtain data
with words and chunk tags only (words outside of a chunk keep their
POS tag).
Again we have evaluated sets of features which hold a single type of
information, words (w--) or chunk tags (c--), or pairs of information,
words and chunk tags (wc--).

\begin{table}[t]
\begin{center}
\begin{tabular}{ r|l|c|c|c|c|r|l|c|c|c|c|}\cline{3-6}\cline{9-12}
\multicolumn{1}{l}{} & \multicolumn{1}{l|}{train} & 
 0 & 1 & 2 & 3 & \multicolumn{1}{l}{~~~~} & train &
 0 & 1 & 2 & 3 \\\cline{2-6}\cline{8-12}
 1 & w         & 61.77 & 84.40 & 83.74 & 81.08 &
 1 & w         & 61.11 & 75.99 & 77.52 & 77.63 \\
 2 & p         & 30.44 & 80.40 & 80.47 & 76.85 &
 2 & p         & 61.71 & 77.52 & 78.74 & 77.95 \\
 3 & c         & 13.67 & 76.76 & 79.05 & 78.71 &
 3 & c         & 00.00 & 67.25 & 75.06 & 75.70 \\
 4 & wp        & 62.24 & 87.19 & 84.45 & 81.22 &
 4 & wp        & 61.25 & 76.52 & 77.92 & 78.12 \\
 5 & wc        & 67.95 & 87.31 & 85.74 & 82.97 &
 5 & wc        & 61.01 & 75.96 & 77.46 & 77.79 \\
 6 & pc        & 49.29 & 86.65 & 84.92 & 81.72 &
 6 & pc        & 61.74 & 77.44 & 78.40 & 77.93 \\
 7 & wpc       & 68.66 & 87.92 & 85.93 & 83.28 &
 7 & wpc       & 61.21 & 76.17 & 77.73 & 78.00 \\\cline{2-6}\cline{8-12}
 8 & 1+2+3     & 38.32 & 85.24 & 86.92 & 85.38 &
 8 & 1+2+3     & 61.67 & 75.93 & 79.60 & 79.94 \\
 9 & 4+5+6     & 68.04 & 88.83 & 87.44 & 84.98 &
 9 & 4+5+6     & 61.44 & 77.30 & 79.15 & 79.38 \\
10 & 7+8+9     & 68.03 & 88.75 & 87.72 & 85.45 &
10 & 7+8+9     & 61.44 & 77.20 & 79.25 & 79.60 \\\cline{2-6}\cline{8-12}
11 & w-        & 54.05 & 83.70 & 83.48 & 81.25 &
11 & w-        & 61.24 & 76.01 & 78.69 & 79.25 \\
12 & c-        & 14.26 & 77.70 & 79.30 & 78.50 &
12 & c-        & 61.73 & 76.82 & 78.34 & 80.90 \\
13 & wc-       & 58.47 & 86.53 & 85.74 & 82.77 &
13 & wc-       & 61.43 & 76.77 & 80.15 & 81.61 \\\cline{2-6}\cline{8-12}
\end{tabular}
\end{center}
\caption{
F$_{\beta=1}$ rates obtained in 10-fold cross-validation experiments
with the training data while predicting open clause brackets (left)
and close clause brackets (right).
We used different combinations of information (w: words, p: POS tags
and c: chunk tags) and different context sizes (0-3).
The best results for open brackets have been obtained with a majority
vote of three information pairs while using context size 1 (row 9)
For close clause brackets best results were obtained with words and
POS tags after compressing the chunks and while using context size 3
(row 13).
} 
\label{tab-cl10}
\end{table}

We have evaluated the twelve groups of feature sets while predicting
the clause open and clause close brackets.
The results can be found in Table \ref{tab-cl10}.
The learner performed best while predicting open clause brackets with
information about the words immediately next to the current word 
(column 1).
When more information was available, its performance dropped slightly.
Of the different feature sets tested, the majority vote of sets that
used pairs of information performed best (column 1, row 9).
The classifiers that generated close brackets improved whenever extra
context information became available.
The best performance was reached while using a pair of words and chunk
tags in the summarized format (column 3, row 13).
We have performed an extra experiment to test if the system improved
when using four context words rather than three.
With words and chunk tags in the summarized format the system obtained 
F$_{\beta=1}$=81.72 for context size four compared with 81.61 for
context size three.
This increase is small so we have chosen context size three for our
further experiments.

With the streams of open and close brackets, we attempted to generate
balanced clause structures by modifying the data streams with a set of
heuristic rules.
In these rules we gave more confidence to the open bracket predictions
since, as can be seen in Table \ref{tab-cl10} the system performs
better in predicting open brackets than close brackets.
After testing different rule sets created by hand and evaluating these
on the available training data, we decided on using the following rule
set:

\begin{enumerate}
\itemsep -0.1cm
\item Assume that exactly one clause starts at each clause start
      position.
\item Assume that exactly one clause ends at each clause end
      position but
\item ignore all clause end positions when currently no clause is
      open, and 
\item ignore all clause ends at non-sentence-final positions
      which attempt to close a clause started at the first word of the
      sentence.
\item If clauses are opened but not closed at the end of the sentence
      then close them at the penultimate word of the sentence.
\end{enumerate}

\noindent
These rules were able to generate complete and consistent embedded
clause structures for the output that the system generated for the
training data of the CoNLL-2001 shared task.
The rules have one main defect: they are incapable of predicting that
two or more clauses start at the same position.
This will make it impossible for the system to detect such clause
start but unfortunately, according to our rule set evaluation, adding
recognition facilities for such multiple clause start would have a
negative influence on overall performance levels.
This set of rules obtained a clause F$_{\beta=1}$ of 71.34 on the
training data of this task when applied to the best results for open
and close brackets.
The rules did not change the open bracket positions and on average the
changes they made to the close bracket positions were an improvement
(F$_{\beta=1}$ = 84.11 compared to 81.61).

An argument which could be made is that since open bracket prediction
is more accurate than close bracket prediction, one could use the
information of the open bracket positions when predicting clause 
close brackets.
We have attempted to do this by repeating the experiment with the best
configuration for close brackets (wc-- with context size 3) while
adding a feature which stated at which clause level the current word
was, according to earlier open and close brackets.
This approach improved the F$_{\beta=1}$ rate of the close bracket
predictor from 81.61 to 83.50.
However, after applying the balancing rules to the open brackets and
the improved close brackets, we only got a clause F$_{\beta=1}$ of
71.39, a minimal improvement over the previous 71.34.
It seems that the extra performance gain obtained in the close bracket
predictor was obtained by solving problems which could already be
solved by the balancing rules.

We applied the balancing rules together with an open bracket predictor
using a combination of pairs of feature types (context size 1) and a
close bracket predictor using summarized pairs of words and chunk tags
(context size 3) to the data files of the CoNLL-2001 shared task.
Our clause identification method obtained an F$_{\beta=1}$ rate of
67.79 for identifying complete clauses (precision 76.91\% and recall
60.61\%). 
In the CoNLL-2001 shared task, the system finished third of six
participants.
One system outperformed the others by a large margin: the boosted
decision tree method by \cite{carreras2001}.
Their system obtained an F$_{\beta=1}$ rate of 78.63 on this task.
The main difference between their approach and ours is that they use a
larger number of features, methods for predicting multiple
co-occurring clause starts and a more advanced statistical model for 
combining brackets to clauses.

In a post-conference study, we have attempted to estimate more 
precisely the cause of the performance difference between our method
an the boosted decision trees used by \cite{carreras2001}.
Our hypothesis was that not only the choice of system made a
difference, but also the choice of features.
For this purpose, Carreras and M\`arques kindly repeated an experiment
in predicting open brackets but this time while using our feature set:
pairs of information using a window of one word left and one right,
while results were combined with majority voting (Table
\ref{tab-cl10}, left, row 9, column 1).
The experiment was performed while testing on the CoNLL-2001
development data set.
Originally the memory-based learner obtained F$_{\beta=1}$ = 89.80 on
this data set while their boosted decision tree approach reached
93.89.
However, while using the memory-based feature set, the performance of
the decision trees dropped to 91.32.
When both systems use the same features, the boosted decision trees
outperform the memory-based learner.
But it is able to perform better with its own feature set.
Our hypothesis was correct: the performance difference between the two
approaches was both caused by choice of the learner and the choice of
the feature set.

The next obvious question is whether the memory-based system would
perform better with the feature set of the boosted decision trees.
Providing an answer to this question was nontrivial.
The feature set consisted of thousands of binary features which were
more than the memory-based learner could handle.
After converting the features from binary-valued to multi-valued,
there were about 70 features left.
At best, the system obtained F$_{\beta=1}$ = 90.52 with this feature
set.
Since we feared that still the number of features was too large for
the system to handle, we performed a forward sequential selection
search process in the feature space starting with zero features.
The memory-based learner reached an optimal performance with 13
features at F$_{\beta=1}$ = 91.82.
These results show that there is still room for improvement for the
memory-based learner but that cooperation with a feature selection
method will be helpful.


\subsection{Noun Phrase Parsing}
\label{sec-npp}

Noun phrase parsing is similar to noun phrase chunking but this time
the goal is to find noun phrases at all levels.
This means that just like in the clause identification task we need to
be able to recognize embedded phrases.
The following example sentence will illustrate this:

\begin{quote}
In ( early trading ) in ( Hong Kong ) ( Monday ) , ( gold ) was quoted \\
at ( ( \$ 366.50 ) ( an ounce ) ) .
\end{quote}

\noindent
This sentence contains seven noun phrases of which the one containing
the final four words of the sentence consists of two embedded noun
phrases.
If we use the same approach as for clause identification, retrieving
brackets of all phrase levels in one step and balancing these, we will
probably not detect this noun phrase because it starts and ends
together with other noun phrases.
Therefore we will use a different approach here.

We will recover noun phrases at different levels by performing
repeated chunking \citep{tks2000naacl}.
We will start with data containing words and part-of-speech tags and
identify the base noun phrases in this data with techniques used in
our noun phrase chunking work.
After this we will replace the phrases that were found by the head
words and their tags.
This will create a summary of the sentences with words and a mixed
data stream of POS tags and chunk tags.
We can apply our noun phrase chunking techniques to this data one more 
time and find noun phrases one level above the base level.
The compressing and chunking steps will be repeated in order to
retrieve phrases at higher levels.
The process will stop when no new phrases are found.

The approach described here seems a trivial expansion of our noun
phrase chunking work.
However, there are some details left to discuss.
First, there is the selection of the head word duing the phrase
summarization process.
At the time we performed these experiments, we did not have access to
the Magerman/Collins set of rules for determining head words, and
therefore we used a rule created by ourselves: the head word of a noun
phrase is the final word of the first noun cluster in the phrase or
the final word of the phrase if it does not contain a noun cluster.

The second fact we should mention, is that the data we used contains
a different format of noun phrase chunks than the data we previously
have worked with.
In this task we use the data set which was developed for the noun
phrase bracketing shared task of CoNLL-99 \citep{osborne99}.
It was extracted from the Wall Street Journal part of the Penn
Treebank \citep{marcus93} without extra modifications and this means,
for example, that possessives between two noun phrases have been
attached to the first one unlike in the noun phrase chunking data.
This and other differences make that we cannot be sure that the
techniques we developed for the other base noun phrase format will
work very well here.
Indeed, there is a performance drop in the chunking part of our shallow
parser when compared with the chunking work (F$_{\beta=1}$ of 92.77
compared with 93.34).
However, we decided not to put extra work in searching for a better
configuration for our noun phrase chunker and have trained an existing
chunker with the data available for this task.

An unforeseen problem occurred when we attempted to use the chunker for
identifying noun phrases above the base level.
Our chunker output is a majority vote of five systems using different
data representations.
In our evaluation work with tuning data (WSJ section 21), we
observed that the overall output of the chunker at nonbase levels was
worse than the performance of the best individual system
\citep{tks2000naacl}. 
The reason for this is that the system that used the O+C data
representation, outperformed the other four systems by a large margin.
Because of this, and probably because the other four systems
made similar errors, the errors of the four cancelled some of the
correct analyses of the best system and caused the majority vote to be
worse than the best individual system.
For this reason we have decided to use only the bracket
representations when processing noun phrases above base levels.

The main open question in this study is what training data to use
when processing the nonbase noun phrases.
In order to find an answer to this question we have tested several
configurations while processing tuning data, WSJ section 21, with
the training data for the CoNLL-99 shared task.
We have tested six training data configurations for predicting open
and close bracket positions: using all bracket positions, those of
base phrases only, those of all phrases except base phrases, those of
phrases of the current level only, those of the current level and the
previous, and those of the current level and the next.
At all levels, using the brackets of the current level only proved to
be working best or close to best.
At the sixth level no new noun phrases were detected.
Therefore we decided to use only brackets of one phrase level in the
training data for nonbase phrases and stop phrase identification after
six levels.

We have applied a noun phrase chunker with fixed symmetrical context
sizes to the noun phrase data of the CoNLL-99 shared task
\citep{tks2000naacl}.
The chunker generated a majority vote of open and close brackets put
forward by five systems, each of which used a different representation
of the base noun phrases (IOB1, IOB2, IOE1, IOE2 and O or C).
All systems used a window of four left and four right for words and
POS tags (18 features) and the four systems using IO representations
additionally performed and extra pass with a window of three left and
three right for words and POS tags, and a window of two left and two
right without the focus tag for chunk tags (also 18 features).
The output of the chunker was presented to a cascade of six chunkers,
each of which consisted of a pair of open and close bracket predictors
which were trained with brackets from one of the levels 1 to 6.
After each chunk phase the phrases found were replaced by the head
word of the phrase and a fixed chunk tag.

The system obtained an overall F$_{\beta=1}$ rate of 83.79 (precision
90.00\% and recall 78.38\%) for identifying arbitrary noun
phrases.\footnote{This performance was already reported by
\cite{tks2000naacl}.}
It is slightly better than our performance at CoNLL-99 (82.98,
obtained without system combination) which was the best of two entries
submitted for the shared task at that workshop.
The performance of our noun phrase chunker can be regarded as a
baseline score for this data set.
This score is already quite high: F$_{\beta=1}$ = 79.70, and it seems
that the nonbase level chunkers have not been contributing much to the
performance of this shallow parser.
Out of curiosity we have also examined how well a full parser does on
the task of identifying arbitrary noun phrases.
For this purpose we looked at output data of a parser described by
\cite{collins99} which was provided with the parser code (WSJ section
23, model 2).
The parser obtained F$_{\beta=1}$ = 89.8 (precision 89.3\% and recall
90.4\%) for this task.
This is a lot better than our shallow parser but we should note that
compared with our application, the Collins parser has access to better
part-of-speech tags and more training data with more sophisticated
annotation rather than only noun phrase boundaries.

\subsection{Full Parsing}
\label{sec-par}

The approach  for parsing noun phrases outlined in the previous
section can be used for generating parse trees containing phrases of
arbitrary phrases as well.
In that case we would be using chunking techniques for performing full 
parsing. 
The is not a new idea.
\cite{ejerhed83} present a Swedish grammar which includes noun phrase
chunk rules.
\cite{abney91} describes a chunk parser which consists of two parts:
one that finds base chunks and another that attaches the chunks
to each other in order to obtain parse trees.
\cite{daelemans95} suggested to find long-distance dependencies with a
cascade of lazy learners among which were constituent identifiers. 
\cite{ratnaparkhi98} built a parser based on a chunker with an
additional bottom-up process which determines at what position to
start new phrases or to join constituents with earlier ones.
With this approach he obtained state-of-the-art parsing results.
\cite{brants99} applied a cascade of Markov model chunkers to the task
of parsing German sentences.
We have extended our noun phrase parsing techniques to parsing
arbitrary phrases \citep{tks2001clin}.
We will present the main findings of this study here as well.

The standard data sets for testing statistical parsers are different
than the ones we used for our earlier work on chunking and shallow
parsing.
The data sets have been extracted from the Wall Street Journal (WSJ)
part of the Penn Treebank \citep{marcus93} as well but they contain
different segments.
The training data consists of sections 02-21 (39,832 sentences) while
section 23 is used as test data (2416 sentences).
The data sets consists of words, and part-of-speech tags which have
been generated by the part-of-speech tagger described by
\cite{ratnaparkhi96}.
In the data the phrase types ADVP and PRT have been collapsed into one
category and during evaluation the positions of punctuation signs in
the parse tree have been ignored.
These adaptations have been done by different authors in order to make
it possible to compare the results of their systems with the first
study that used these data sets \citep{magerman95} and all follow-up
work.

In our work on arbitrary parsing, we were interested in finding an
answer to four questions.
In order to obtain these answers, we have performed tests with smaller
data sets which were taken from the standard training data for this
task: WSJ sections 15-18 as training data and section 20 as test data.
The first topic we were interested in, was the influence of context
size and size of the examined nearest neighborhood size (parameter k
of the memory-based learner) on the performance of the parser.
We took the noun phrase parser developed in the previous section,
lifted its restriction of generating noun phrases only and applied it
to this data set while using different context sizes and values for
parameter k for the classifiers that identified phrases above the base
levels.
The different types of the chunks were derived by using the
Double-Phase Approach for chunking (see Section \ref{sec-chuarb}).
The best configuration we found was a context of two left and two
right words and POS tags with k is 1.
The nearest neighborhood size is smaller than used in our earlier work
(3) and the best context size is smaller than in our noun phrase
chunking work (4).
However, the best context size we found for this task is exactly the
same as reported by \cite{ratnaparkhi98}.

The second topic we were interested in was the type of training data
that should be used for finding phrases above the base level.
In our noun phrase parsing work, we found that the best performance
could be obtained by using only data of the current phrase level.
This will cause a problems for our parser, since the tree depth may
become as large as 31 in our corpus but there will be few training
material available for these high level phrases if we use the same
training configuration as in our noun phrase parsing work.
We have tested two different training configurations to see if we
could use more training data for this task without losing performance.
With the first of these, using the current, previous and next phrase level,
performance was as well (F$_{\beta=1}$=77.13) as while using only the
current level (77.17).
However, when we trained the cascade of chunkers while using brackets
of all phrase levels, the performance dropped to 67.49.
We have decided to keep on using the current phrase level only in the
training data despite its problems with identifying higher level
phrases. 

In the results that we have presented in this paper, the precision
rates have always been higher than the recall rates.
For a part, this is caused by the method we use for balancing open
brackets and close brackets.
It removes all brackets which cannot be matched with another one which
is approximately the same as accepting clauses which are likely to be
correct and throwing away all others.
We wanted to test if we could obtain more balanced precision and
recall rates because we hoped that these would lead to a better
F$_{\beta=1}$ rate.
Therefore we have tested two alternative methods for combining
brackets.
The first disregarded the type of the open brackets and allowed close
brackets to be combined with open brackets of any type.
The second method allowed open brackets to match with close brackets
of any type.
Unfortunately neither the first (F$_{\beta=1}$=72.33) nor the second
method (76.06) managed to obtain the same F$_{\beta=1}$ rate as our
standard method for combining brackets.
Therefore we decided to stick with the latter.

The final issue which we wanted to examine is the performance
progression of the parser at the different levels of the process.
The recall of the parser should increase for every extra step in the
cascade of chunkers but we would also like to know how precision and
F$_{\beta=1}$ progressed.
We have measured this for our small parameter tuning data set and
found that indeed recall increased until level 30 of a maximum of 32
and remained constant after that.
Precision dropped until the same level, remaining at the same value
afterwards while F$_{\beta=1}$ reached a top value at level 19 and
dropped afterwards.
The reason for the later drop in F$_{\beta=1}$ value is that while
the recall is still rising, it cannot make up for the loss of
precision at later levels.
Since we want to optimize the F$_{\beta=1}$ rate, we have decided to
restrict the number of cascaded chunkers in our parser to 19 levels.
We have added an extra post-processing step which after the 19 levels
of processing adds clause brackets (S) around sentences which have not 
already been identified as a clauses.

We have applied the best parser configuration found to the standard
parsing data.
Our parser used an arbitrary chunker with the configuration described
in Section \ref{sec-chuarb} (a Majority Vote of five systems using
different data representations) but trained with the relevant data for
this task.
Higher level phrases were identified by a cascade of 19 chunkers, each
of which had a pair of independent open and close bracket classifiers
which used a context of two left and two right of words and POS tags
while being trained with brackets of the current level only.
At each level, open and close brackets were combined to chunks by
removing all brackets that could not be matched with a bracket of the
same type.
The parser contained a post-processing process which added clause
brackets around sentences which were not identified as a clause after
the 19 processing stages.
This chunk parser obtained an F$_{\beta=1}$ rate of 80.49 on WSJ
section 23 (precision 82.34\% and recall 78.72\%).

The performance of our chunk parser is modest compared with
state-of-the-art statistical parsers, which obtain around 90
F$_{\beta=1}$ rate \citep{collins99,charniak2000}.
However, we have a couple of suggestions for improving its
performance.
First, we could attempt giving the parser access to more information,
for example about lower phrase levels.
Currently, the parser only knows the head words and phrase types of
daughters of phrases that are being built and this might not be
enough. 
Second, we could try to find a better method for predicting bracket
positions.
For reasons explained in the previous section, we could not use a
majority vote of systems using different representations.
This might have helped to obtain a better performance.
Finally, we would like to change the greedy approach of our parser.
Currently it chooses the best segmentation of chunks at each level and
builds on that but ideally it would be able to remember some
next-to-best configurations as well and perform backtracking from the
earlier choices whenever necessary.
This approach would probably improve performance considerably
(as shown by \cite{ratnaparkhi98}, Table 6.5).
A practical problem which needs to be solved here, is that in nearest
neighbor memory-based learning alternative classes do not receive
confidence measures.
Rather, sets of item-dependent distances are used to determine the
usability of the classes.
Comparing partial trees requires comparing sets of distances and
it is not obvious how this should be done.

These extra measures will probably improve the performance of the
chunk parser.
However, it is questionable whether it is worthwhile continuing with this
approach.
The present version of the parser already requires a lot of memory and
processing time: more than a second per {\it word} for chunking only
compared with a mere 0.14 seconds per {\it sentence} for a statistical
parser which performed better \citep{ratnaparkhi98}.
Extra extensions will probably slow down the parser even more so we
are not sure if extending this approach is worth the trouble.

\begin{table}[t]
\begin{center}
\begin{tabular}{|l|c|c|c|}\cline{2-4}
\multicolumn{1}{l|}{section 20} & 
   precision & recall & F$_{\beta=1}$ \\\hline
\cite{kudoh2001}     & 94.15\% & 94.29\% & 94.22 \\
\cite{tks2000coling} & 94.18\% & 93.55\% & 93.86 \\
MBL                  & 94.01\% & 92.67\% & 93.34 \\
\cite{tks2000naacl}  & 93.63\% & 92.89\% & 93.26 \\
\cite{munoz99}       & 92.4\%  & 93.1\%  & 92.8  \\
\cite{ramshaw95}     & 91.80\% & 92.27\% & 92.03 \\
\cite{argamon99}     & 91.6\%  & 91.6\%  & 91.6  \\\hline
baseline             & 78.20\% & 81.87\% & 79.99 \\\hline
\end{tabular}
\end{center}
\caption{A selection of results that have been published for the
Ramshaw and Marcus data sets for noun phrase chunking.
Our chunker (MBL) is third-best.
The baseline results have been produced by a system that selects the
most frequent chunk tag (IOB1) for each part-of-speech tag.
The best performance for this task has been obtained by a system using
Support Vector Machines 
\citep{kudoh2001}.
}
\label{tab-resnp}
\end{table}

\section{Related Work}

In this section we will compare our work with that of others that have
applied machine learning techniques to the same data sets.
First we will discuss the two chunking tasks and then the tasks that
required output of embedded structures.
Many systems have been applied to the five tasks.
Rather that giving a detailed description of each of them, we will
list the best performing systems for each task and mention some 
differences between these systems and ours.
This comparison of our memory-based shallow parsers with other work
shows that they produce state-of-the-art results for the chunking
tasks but not for the tasks which require identification of embedded
structures.

\subsection{Chunking}

Table \ref{tab-resnp} shows a selection of the best results published
for the noun phrase chunking task.\footnote{An elaborate overview of
most of the systems that have been applied to this task can be found
on http://lcg-www.uia.ac.be/\~{ }erikt/research/np-chunking.html}
As far as we know, the results presented in this paper (line MBL) are
the third-best results.
We have participated in producing the second-best result
\citep{tks2000coling} which was produced by combining of the results
of five different learning techniques.
The best results for this data set have been generated with Support
Vector Machines \citep{kudoh2001}.\footnote{Although we do not wish to
underestimate the power of Support Vector Machines, we should note
that it seems that the optimal results presented by \cite{kudoh2001}
have been obtained by tuning the system to the test data.}
A statistical analysis of our current result revealed that all
performances outside of the region 92.93-93.73 are significantly
different from ours.
This means that all results in the table, except from the 93.26, are
significantly different from ours. 

A topic to which we have paid little attention is the analysis of the
errors that our approach makes.
Such an analysis would provide insights into the weaknesses of the
system and might provide clues to methods for improving the system.
For noun phrase chunking we have performed a limited error analysis
by manually evaluating the errors that were made in the first section
of a 10-fold cross-validation experiment on the training data while
using the chunker described by \cite{tks2000naacl}.
This analysis revealed that the majority of the errors were caused by
errors in the part-of-speech tags (28\% of the false positives/29\% of
the false negatives).
In order to acquire reasonable results, it is custom not to use the
part-of-speech tags from the Treebank, but use tags that have been
generated by a part-of-speech tagger.
This prevents the system performance from reaching levels which would
be unattainable for texts for which no perfect part-of-speech tags
exist.
Unfortunately the tagger makes errors and some of these errors cause
the noun phrase segmentation to become incorrect.

The second most frequently occurring error cause was related to
conjunctions of noun phrases (16\%/18\%).
Deciding whether a phrase like {\it red dwarfs and giants} consist of
one or two noun phrases requires semantic knowledge and might be too
ambitious for present-day systems to solve.
The other major causes of errors all relate to similar hard cases:
attachment of punctuation signs (15\%/12\%; inconsistent in the
Treebank), deciding whether ambiguous phrases without conjunctions
should be one or two noun phrases (11\%/12\%), adverb attachment
(5\%/4\%), noun phrases containing the word {\it to} (3\%/3\%),
Treebank noun phrase segmentation errors (3\%/1\%) and noun phrases
consisting of the word {\it that} (0\%/2\%).
Apart from these hard cases there also were quite a few errors for
which we could not determine an obvious cause (19\%/19\%).

The most obvious suggestion for improvement that came out of the error
analysis was to use a better part-of-speech tagger.
We are currently using the Brill tagger \citep{brill94}.
Better taggers are available nowadays but using the Brill tags here 
was necessary in order to be able to compare our approach with earlier
studies, which have used the Brill tags as well. 
The error analysis did not produce other immediate suggestions for
improving our noun phrase chunking approach.
We are relieved about this because it would have been an
embarrassment if our chunker had produced systematic errors.
However, there is a trivial way to improve the results of the noun
phrase chunker: by using more training data.
Different studies have shown that by increasing the training data size
by 300\%, the F$_{\beta=1}$ error might drop with as much as 25\%
\citep{ramshaw95,tks2000naacl,kudoh2001}.
Another study for a different problem, confusion set disambiguation,
has shown that a further cut in the error rate is possible with even
larger training data sets \citep{banko2001}.
In order to test this for noun phrase chunking we need a hand-parsed
corpus which is larger than anything that is presently available.

\begin{table}[t]
\begin{center}
\begin{tabular}{|l|c|c|c|}\cline{2-4}
\multicolumn{1}{l|}{section 20} & 
   precision & recall & F$_{\beta=1}$ \\\hline
\cite{zhang2001}     & 94.29\% & 94.01\% & 94.13 \\
\cite{kudoh2001}     & 93.89\% & 93.92\% & 93.91 \\
\cite{kudoh2000}     & 93.45\% & 93.51\% & 93.48 \\
\cite{hvh2000}       & 93.13\% & 93.51\% & 93.32 \\
\cite{tks2000conll}  & 94.04\% & 91.00\% & 92.50 \\
\cite{zhou2000}      & 91.99\% & 92.25\% & 92.12 \\
\cite{dejean2000}    & 91.87\% & 92.31\% & 92.09 \\\hline
baseline             & 72.58\% & 82.14\% & 77.07 \\\hline
\end{tabular}
\end{center}
\caption{A selection of results that have been published for the
arbitrary chunking data set of the CoNLL-2000 shared task.
Our chunker 
\citep{tks2000conll} 
is fifth-best.
The baseline results have been produced by a system that selects the
most frequent chunk tag (IOB1) for each part-of-speech tag.
The best performance for this task has been obtained by a system using
regularized Winnow 
\citep{zhang2001}.
Systems that have been applied both to the arbitrary chunking task and
the noun phrase chunking task performed approximately equally well for
NP chunks in both tasks.
}
\label{tab-resxp}
\end{table}

Table \ref{tab-resxp} contains a selection of the best results published
for the arbitrary chunking data used in the CoNLL-2000 shared
task.\footnote{More results for the chunking task can be found on
http://lcg-www.uia.ac.be/conll2000/chunking/}
Our chunker \citep{tks2000conll} is the fifth-best on this list.
Immediately obvious is the imbalance between precision and recall: 
the system identifies a small number of phrases with a high precision
rate.
We assume that this is primarily caused by our method for generating
balanced structures from streams of open and close brackets.
We have performed a bootstrap resampling test on the chunk tag
sequence associated with this result.
An evaluation of 10,000 pairs indicated that the significance interval
for our system (F$_{\beta=1}$ = 92.50) is 92.18-92.81 which means that
all systems that are ahead of ours perform significantly better and all
systems that are behind perform  significantly worse.
We are not sure what is causing these large performance differences.
At this moment we assume that our approach has difficulty with
classification tasks when the number of different output classes
increases.

\begin{table}[t]
\begin{center}
\begin{tabular}{|l|c|c|c|}\cline{2-4}
\multicolumn{1}{l|}{section 21} 
                      & precision & recall & F$_{\beta=1}$ \\\hline
\cite{carreras2001}   & 84.82\% & 73.28\% & 78.63 \\
\cite{molina2001}     & 70.89\% & 65.57\% & 68.12 \\
\cite{tks2001conll}   & 76.91\% & 60.61\% & 67.79 \\
\cite{patrick2001}    & 73.75\% & 60.00\% & 66.17 \\
\cite{dejean2001}     & 72.56\% & 54.55\% & 62.77 \\
\cite{hammerton2001b} & 55.81\% & 45.99\% & 50.42 \\\hline
baseline              & 98.44\% & 31.48\% & 47.71 \\\hline
\end{tabular}
\end{center}
\caption{Results of the clause identification part of the CoNLL-2001
shared task.
Our clause identifier \citep{tks2001conll} is third-best.
The baseline results have been produced by a system that only puts
clause brackets around complete sentences.
The best performance for this task has been obtained by a system using
boosted decision trees \citep{carreras2001}.
}
\label{tab-rescl}
\end{table}

\subsection{Parsing}

A complete overview of the clause identification results of the
CoNLL-2001 shared task can be found in Table \ref{tab-rescl}
\citep{tksdejean2001conll}.
Our approach was third-best.
A bootstrap resampling test with a population of 10,000 random samples
generated from our results produced the 90\% significance interval
66.66-68.95 for our system which means that our result is not 
significantly different from the second result.
The boosted decision trees used by \cite{carreras2001} did a lot
better than the other systems.
In Section \ref{sec-clauses}, we have made a comparison between the
performance of this system and ours and concluded that the performance
differences were both caused by the choice of learning system and
by a difference in the features chosen for representing the task. 

The noun phrase parsing task has not received much attention in the
research community and there are only few results to compare with.
\cite{osborne99} used a grammar-extension method based on Minimal
Description Length and applied it to a Definite Clause Grammar.
His system used different training and test segments of the Penn
Treebank than we did.
At best, it obtained an F$_{\beta=1}$ rate of 60.0 on the test data
(precision 53.2\% and recall 68.7\%).
\cite{krymolowski2000} applied a memory-based learning technique
specialized for learning sequences to a noun phrase parsing task.
Their system obtained F$_{\beta=1}$=83.7 (precision 88.5\% and recall
79.3\%) on yet another segment of the Treebank.
This performance is very close to that of our approach
(F$_{\beta=1}$=83.79).
The memory-based sequence learner used much more training data than
ours (about four times as much) but unlike our method, it generated
its output without using lexical information, which is impressive.
The performance of the Collins parser on the subtask of noun phrase
parsing which we mentioned in Section \ref{sec-npp}
(F$_{\beta=1}$=89.8) shows that there is room for improvement left for
all systems that were discussed here.\footnote{Our full parser, which
was trained and tested on the same data as the Collins parser,
obtained F$_{\beta=1}$=86.96 for recognizing NP phrases only.}

\begin{table}[t]
\begin{center}
\begin{tabular}{|l|c|c|c|}\cline{2-4}
\multicolumn{1}{l|}{section 23} & 
   precision & recall & F$_{\beta=1}$ \\\hline
\cite{collins2000}   & 89.9\% & 89.6\% & 89.7 \\
\cite{bod2001}       & 89.7\% & 89.7\% & 89.7 \\
\cite{charniak2000}  & 89.5\% & 89.6\% & 89.5 \\
\cite{collins99}     & 88.3\% & 88.1\% & 88.2 \\
\cite{ratnaparkhi98} & 87.5\% & 86.3\% & 86.9 \\
\cite{charniak97}    & 86.6\% & 86.7\% & 86.6 \\
\cite{magerman95}    & 84.3\% & 84.0\% & 84.1 \\
\cite{tks2001clin}   & 82.3\% & 78.7\% & 80.5 \\\hline
\end{tabular}
\end{center}
\caption{A selection of results that have been published for 
parsing sentences shorter than 100 words of the Penn Treebank.
The performance of our parser \citep{tks2001clin} is not quite
state-of-the-art.
The best performance for this task has been obtained by statistical
parsers and data-oriented parsers 
\citep{collins2000,charniak2000,bod2000}.
}
\label{tab-respa}
\end{table}

A selection of results for parsing the Penn Treebank can be found in
Table \ref{tab-respa}.
The F$_{\beta=1}$ error rate of the best systems is about half of that
of ours.
A more detailed comparison of the output data of our memory-based
parser and one of the versions of the Collins parser
\citep[][model 2]{collins99} has shown the large performance difference
is caused by the way nonbase phrases are processed
\citep{tks2001clin}. 
Our chunker performs reasonably well compared with the first stage of
the Collins parser (F$_{\beta=1}$ = 49.30 compared with 49.85).
Especially at the first few levels after the base levels, our parser
looses F$_{\beta=1}$ points compared with the Collins parser.
The initial difference of 0.65 at the base level grows to 2.92 after
three more levels, 5.16 after six and 6.13 after nine levels with a
final difference of 6.59 after 20 levels \citep{tks2001clin}.
At the end of Section \ref{sec-par}, we have put forward some
suggestions for improving our parser.
However, we have also noted that further improvement might not
be worthwhile because it will make our parser even slower than it
already is.

\section{Concluding Remarks}

We have presented memory-based approaches to shallow parsing and we
have applied these to five tasks: noun phrase chunking, arbitrary
chunking, clause identification, noun phrase parsing and full
parsing.
We have used two additional techniques for improving the performance
of our shallow parsers: feature selection and system combination.
The first was used to compensate for a problem of the memory-based
learner: it has difficulty with ignoring features that are not
immediately relevant. 
While feature selection worked well in one study (clause
identification with large feature sets), it did not make much
difference to the overall performance of our noun phrase chunker.
We believe that other techniques that were incorporated in the chunker
(cascading and system combination) have already stretched the
performance of the system to its limits.
Therefore there might not have been much left to gain by using feature
selection.
System combination has proved to be quite useful for generating base
phrases.
Unfortunately, we could not apply it for higher level chunks because 
our method for producing different system results, using different
data representations, failed to produce results for higher level phrases
that could be improved with the Majority Voting technique we used for
chunking.

A comparison of our work with other studies revealed that our
approach works well for base phrase identification, but not for
finding embedded structures.
We have made a couple of suggestions for improving the performance on
tasks that require generating embedded structures: provide different
features to the learners, try to find a method which allows
combination of different systems when working on higher level phrases
and replace the greedy phrase selection approach currently used by one
that allows backtracking from earlier choices.
However, while further improvement is interesting from a scientific
point of view, it might not be useful from a practical point of view.
Our present method is already slower than state-of-the-art full
parsers and it requires more memory.
Extra improvements to this approach will probably slow it down even
more without guaranteeing state-of-the-art performance.

\acks{
\hspace*{-0.3cm}
We would like to thank 
our colleagues of CNTS - Language Technology Group, University of
Antwerp, Belgium and  
ILK, University of Tilburg, The Netherlands, 
the members of the TMR-LCG network, in particular James Hammerton, and
two anonymous reviewers for
valuable discussions and comments. 
We are grateful to Xavier Carreras for his cooperation in the
comparison study of his clause identification system with ours.
This study was funded by the European Training and Mobility of
Researchers (TMR) network Learning Computational 
Grammars.\footnote{http://lcg-www.uia.ac.be/}
}

\vskip 0.2in
\bibliography{ref}

\end{document}